\def\year{2021}\relax
\documentclass[letterpaper]{article} 
\usepackage{aaai21}  
\usepackage{times}  
\usepackage{helvet} 
\usepackage{courier}  
\usepackage[hyphens]{url}  
\usepackage{graphicx} 

\usepackage[hidelinks,colorlinks, linkcolor=red, citecolor=black]{hyperref}
\usepackage{algorithm}
\usepackage{algorithmic}
\usepackage{multirow}
\usepackage{verbatim}
\usepackage{color}
\usepackage{gensymb}
\usepackage{amsmath}
\usepackage{subfiles}
\usepackage{amssymb}  
\newcommand{\tabincell}[2]{\begin{tabular}{@{}#1@{}}#2\end{tabular}}
\usepackage[switch]{lineno}

\usepackage{natbib}  
\usepackage{caption} 
\title{CIA-SSD: Confident IoU-Aware Single-Stage Object Detector From Point Cloud\\}
\author{Wu Zheng, Weiliang Tang, Sijin Chen, Li Jiang, Chi-Wing Fu\\}
\affiliations {
The Chinese University of Hong Kong, China \\
\{wuzheng, lijiang, cwfu\}@cse.cuhk.edu.hk, \{tangwl123, chensjvin\}@foxmail.com }

\begin{document}
\newcommand{\TODO}[1]{{\color{red}{[TODO: #1]}}}
\newcommand{\phil}[1]{{\color[rgb]{0.3,0.7,0.3}{[PH: #1]}}}
\newcommand{\zw}[1]{{\color[rgb]{0.7,0.3,0.7}{[ZW: #1]}}}
\newcommand{\para}[1]{\vspace{.05in}\noindent\textbf{#1}}
\def\ie{\emph{i.e.}}
\def\eg{\emph{e.g.}}
\def\etal{{\em et al.}}
\def\etc{{\em etc.}}

\maketitle

\ifx\allfiles\undefined
\documentclass[letterpaper]{article}
\begin{document}
\else
\chapter{abstraction}
\fi

\begin{abstract}
Existing single-stage detectors for locating objects in point clouds often treat object localization and category classification as separate tasks, so the localization accuracy and classification confidence may not well align.
To address this issue, we present a new single-stage detector named the {\em Confident IoU-Aware Single-Stage object Detector\/} (CIA-SSD).
First, we design the lightweight {\em Spatial-Semantic Feature Aggregation\/} module to adaptively fuse high-level abstract semantic features and low-level spatial features for accurate predictions of bounding boxes and classification confidence.
Also, the predicted confidence is further rectified with our designed {\em IoU-aware confidence rectification module\/} to make the confidence more consistent with the localization accuracy.
Based on the rectified confidence, we further formulate the {\em Distance-variant IoU-weighted NMS\/} to obtain smoother regressions and avoid redundant predictions.
We experiment CIA-SSD on 3D car detection in the KITTI test set and show that it attains top performance in terms of the official ranking metric (moderate AP $80.28\%$) and above 32 FPS inference speed, outperforming all prior single-stage detectors.
The code is available at \url{https://github.com/Vegeta2020/CIA-SSD}.
\end{abstract}

\ifx\allfiles\undefined
\end{document}
\fi 
\ifx\allfiles\undefined
\documentclass[letterpaper]{article}
\begin{document}
\else
\chapter{introduction}
\fi

\section{1\quad Introduction}
To detect objects in autonomous driving, point clouds are often adopted to offer robust information.
In general, there are two classes of methods to detect objects in point clouds: single-stage and two-stage.
Though two-stage detectors usually attain higher average precisions benefited from an extra refinement stage, single-stage detectors are typically more efficient due to their simpler network structures.
Also, the detection precisions of recent single-stage detectors~\cite{he2020structure,yang20203dssd,shi2020point} gradually approach that of the state-of-the-art two-stage detectors.
The advantages of time efficiency and competitive precision motivate us to focus this work on single-stage detectors.

\begin{figure}[!t]
\centering
\hspace*{-8mm}
\includegraphics[width=0.96\columnwidth]{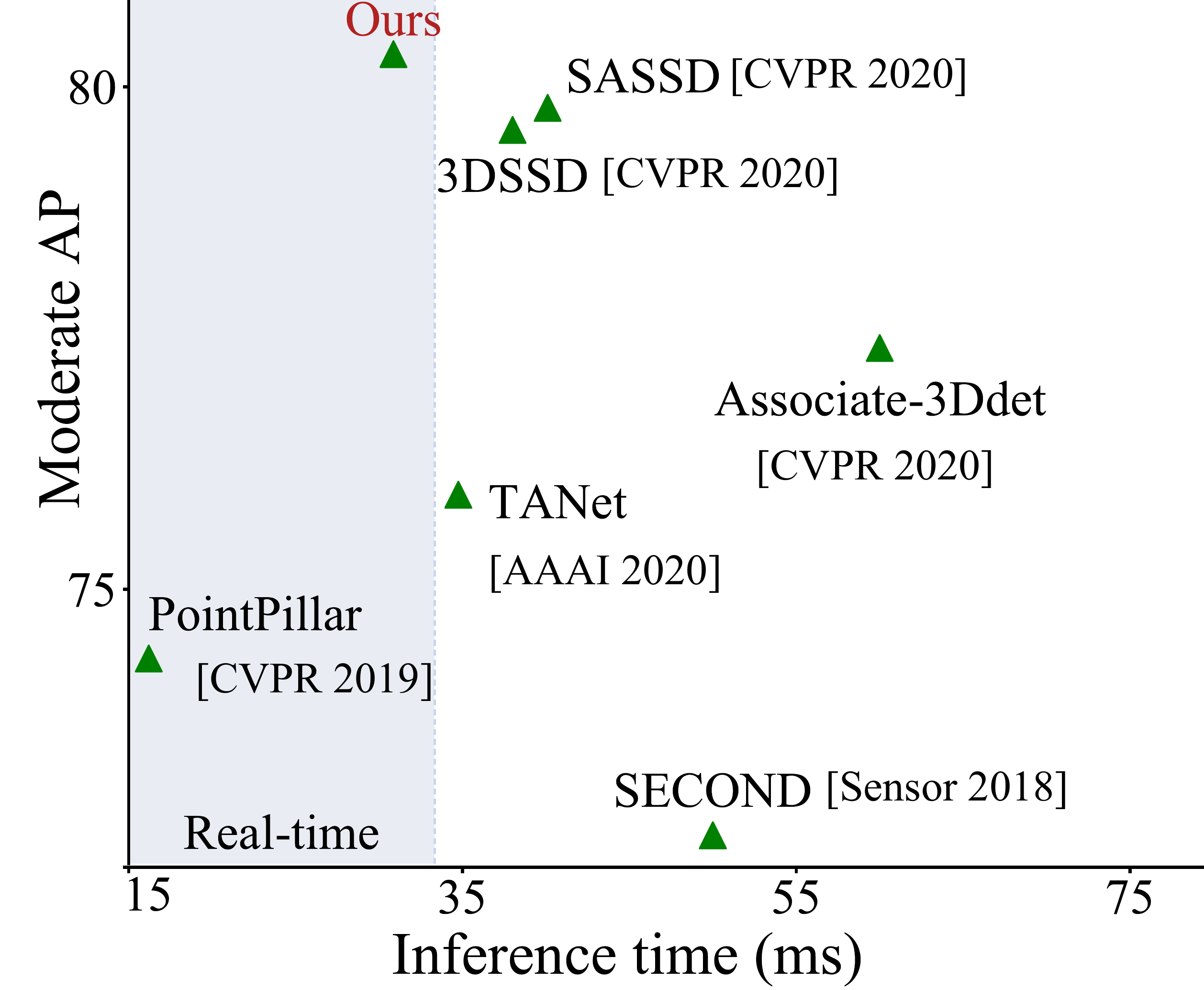}
\vspace*{-2mm}
\caption{Our CIA-SSD attains top performance (official rank: moderate AP $80.28\%$) and real-time speed ($30.76$ ms) on 3D car detection in KITTI test set~\protect\cite{geiger2013vision}, compared with the state-of-the-art single-stage detectors.}
\label{fig:AP_Time}
\vspace*{-3mm}
\end{figure}

Existing 3D object detectors often treat object localization and category classification as separate tasks, so the localization accuracy and classification confidence may not align well~\cite{jiang2018acquisition}.
Hence, two-stage detectors~\cite{yang2019std,shi2020pv} extract features from the region proposals generated by the first-stage backbone and predict the IoUs between the regressed bounding boxes and ground-truth boxes in the second stage to refine the confidence predictions.
Compared with hard-category labels, the soft IoU labels are usually more consistent with the localization qualities, thus leading to more accurate confidence predictions.

Compared with two-stage detectors, single-stage detectors cannot train features extracted from their predicted bounding boxes with a second-stage network.
Also, their features are learned mostly based on the pre-defined anchors or classified positive points, so the resulting IoU predictions may not be as accurate as those in the two-stage networks.
Hence, general single-stage detectors cannot effectively rectify confidence predictions like the two-stage ones.

To resolve this issue, SASSD~\cite{he2020structure}, a very recent single-stage detector, exploits an interpolation approach to obtain the region proposal features for confidence rectification.
Their approach is, however, very complex with the interpolation operation.
In this work, we design a new {\em confidence rectification module\/} embedded in our {\em Confident IoU-Aware Single-Stage object Detector\/} (CIA-SSD) to address the issue more elegantly.
Our key idea is based on the finding that anchor-feature-based IoU predictions are discriminative especially between the precise and imprecise regressions of the bounding boxes.
Thus, by utilizing a convex function to augment the discrimination, we polarize the effect of IoU predictions between the precise and imprecise regressions and effectively rectify the confidence in the post processing.

Besides, we design a lightweight {\em Spatial-Semantic Feature Aggregation\/} (SSFA) module to adaptively fuse high-level abstract semantic features and low-level spatial features for more accurate predictions of bounding boxes and classification confidence.
Compared with the commonly-used 2D feature extraction module~\cite{yan2018second,he2020structure}, our SSFA module boosts the performance effectively with a moderate increase in the model complexity.
Further, we notice that there are often more redundant false-positive predictions and strong oscillations for bounding boxes regressed at large distances from the viewpoint.
Hence, we formulate a novel {\em Distance-variant IoU-weighted NMS\/} (DI-NMS) to obtain smoother regressions and reduce redundant predictions, by considering the depth factor not encountered in 2D NMS.

Figure~\ref{cover_pic} illustrates the quality of our results with an example, in which our method can find better-aligned bounding boxes and avoid redundant predictions, compared with the state-of-the-art single-stage detector SASSD.
Below, we summarize the contributions of this work:
(i) a lightweight spatial-semantic feature aggregation module that adaptively fuses high-level abstract semantic features and low-level spatial features for more accurate bounding box regression and classification;
(ii) an IoU-aware confidence rectification module to alleviate the misalignment between the localization accuracy and classification confidence; and
(iii) a distance-variant IoU-weighted NMS to smooth the regressions and avoid redundant predictions.
CIA-SSD attains top performance (moderate AP $80.28\%$) and real-time speed ($30.76$ ms) on 3D car detection in KITTI test set~\cite{geiger2013vision}, as illustrated in Figure~\ref{fig:AP_Time}.

\ifx\allfiles\undefined
\end{document}
\fi 
\ifx\allfiles\undefined
\documentclass[letterpaper]{article}
\begin{document}
\else
\chapter{related_work}
\fi

\section{2\quad Related Work}
There are two types of LiDAR-based 3D object detectors:
(i) Two-stage detectors usually generate region proposals in the first stage, then feed these regions of interests into a second-stage network for refinement; and
(ii) In contrast, single-stage detectors have simpler networks, since they directly regress class scores and bounding boxes in one stage.

\begin{figure}[!t]
\centering
\includegraphics[width=0.85\columnwidth]{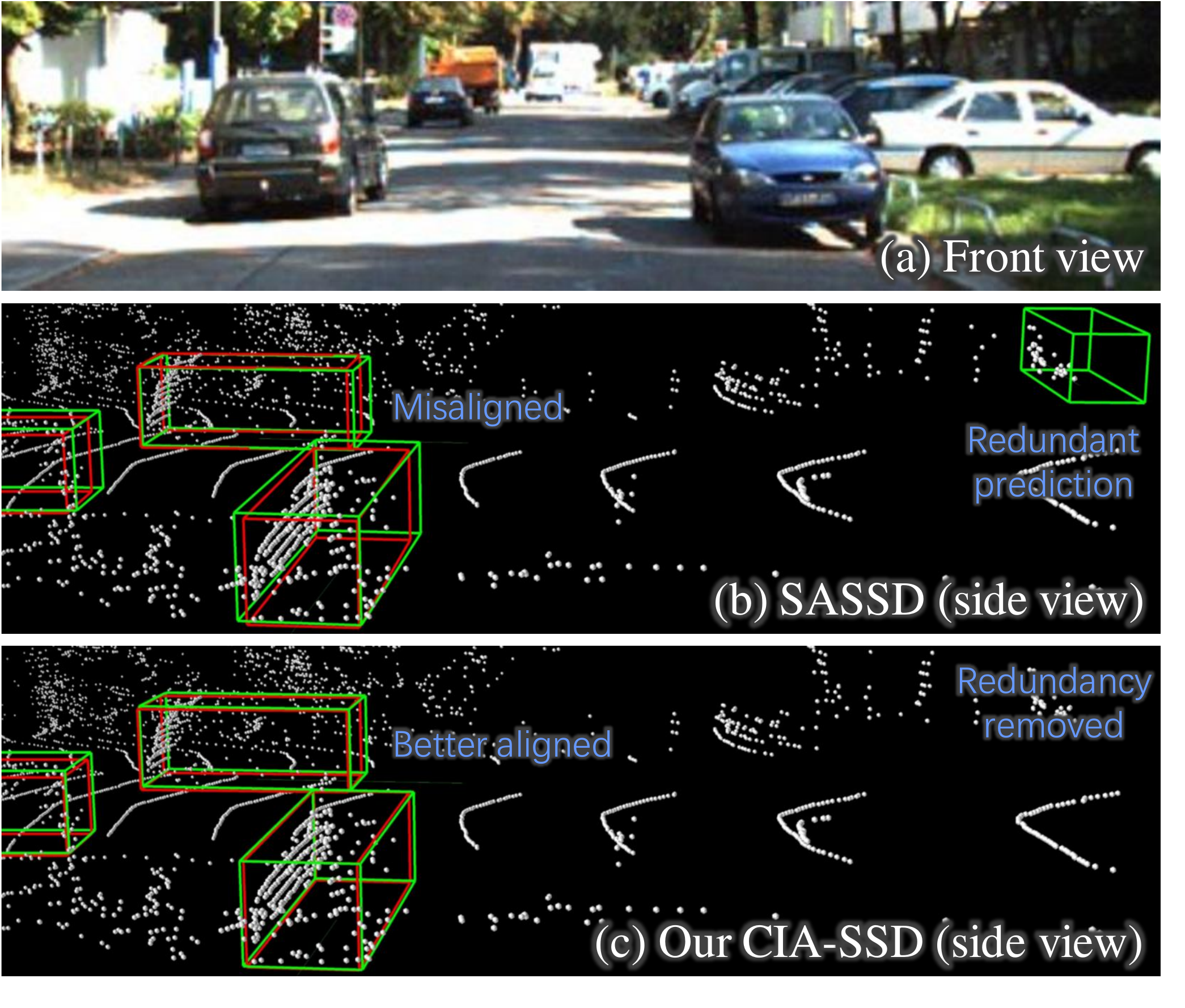}
\vspace*{-2mm}
\caption{As shown in this example, our CIA-SSD (c) can predict bounding boxes (green) that better align with the ground truths (red) and avoid redundant predictions, as compared with the very recent single-stage detector SASSD (b).}
\label{cover_pic}
\vspace*{-3mm}
\end{figure}

Among the two-stage detectors, PointRCNN~\cite{shi2019pointrcnn} uses PointNet++~\cite{qi2017pointnet++} as the backbone in both stages and devises an anchor-free strategy to generate 3D proposals.
Based on PointRCNN, Part-$A^2$~\cite{shi2020points} replaces the PointNet++ backbone with a sparse convolutional network, and proposes RoI-aware point cloud feature pooling in the refinement.
STD~\cite{yang2019std} exploits spherical anchors to generate proposals, together with a segmentation branch, to reduce the number of positive anchors.
PV-RCNN~\cite{shi2020pv} uses set abstraction modules to extract point features from multi-scale voxel features in the first stage to refine the region proposals.

Among the single-stage detectors, VoxelNet~\cite{zhou2018voxelnet} voxelizes the point cloud and proposes a voxel feature encoding layer to extract point features in each voxel and produce fixed-length features for batch training.
PointPillar~\cite{lang2019pointpillars} divides a point cloud into pillars instead of voxels for feature extraction, then utilizes a 2D convolutional detection architecture for object localization.
SECOND~\cite{yan2018second} exploits the sparse convolution~\cite{liu2015sparse} and submanifold sparse convolution~\cite{graham20183d} to replace conventional 3D convolution.
TANet~\cite{liu2020tanet} presents a triple attention module embedded in voxel feature extraction and combines it with a proposed cascaded refinement network.
Very recently, Point-GNN~\cite{shi2020point} proposes a graph neural network to extract point features and achieves a decent performance.
3DSSD~\cite{yang20203dssd} proposes a fusion sampling strategy by combining both feature- and point-based farthest point sampling for better classification performance.
SASSD~\cite{he2020structure} presents an auxiliary network in parallel with a sparse convolutional network to regress the box centers and semantic classes for each point with interpolated voxel features.

As we shall show in the results, recent single-stage detectors have achieved comparable performance (average precision) with the state-of-the-art two-stage detectors.
Given their high efficiency, single-stage detectors have great potential for real-time applications.
This motivates us to focus this research on developing a new single-stage detector CIA-SSD, which has attained top performance and real-time speed, compared with all previous single-stage detectors.

\begin{figure*}
\centering
\includegraphics[width=17.0cm]{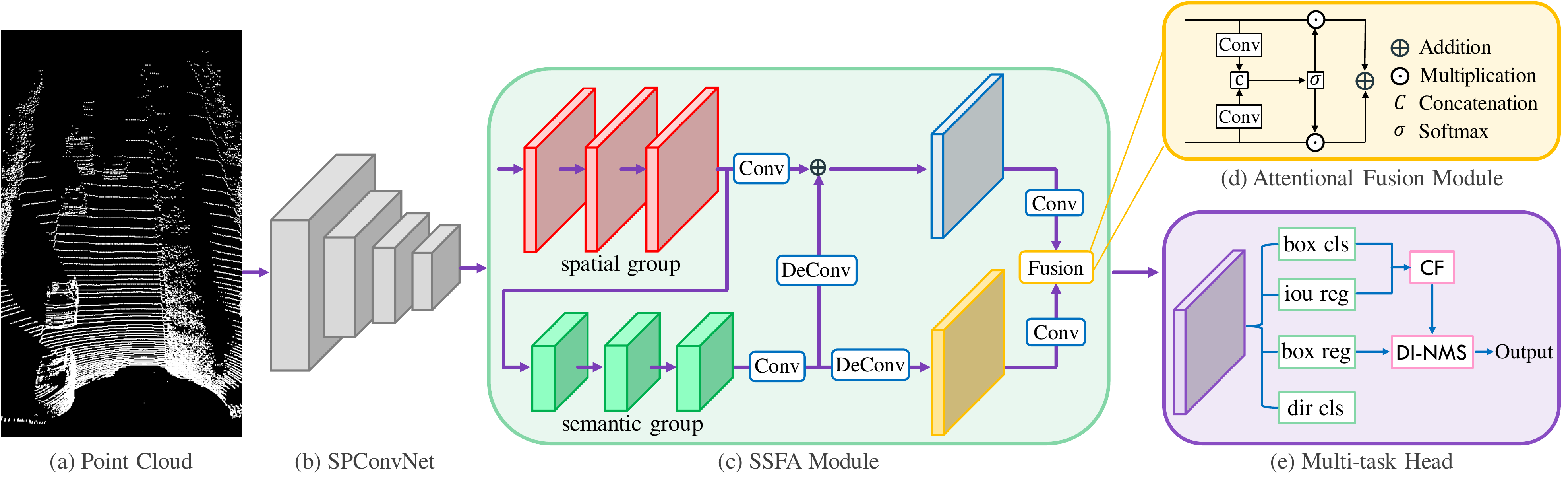} %
\vspace*{-2.5mm}
\caption{The pipeline of our proposed Confident IoU-Aware Single-Stage object Detector (CIA-SSD).
First, we encode the input point cloud (a) with a sparse convolutional network denoted by SPConvNet (b), followed by our spatial-semantic feature aggregation (SSFA) module (c) for robust feature extraction, in which an attentional fusion module (d) is adopted to adaptively fuse the spatial and semantic features.
Then, the multi-task head (e) realizes the object classification and localization, with our introduced confidence function (CF) for confidence rectification. In the end, we further formulate the distance-variant IoU-weighted NMS (DI-NMS) for post-processing.
Note that ``box cls,'' ``iou reg,'' ``box reg,'' and ``dir cls'' in (e) denote bounding box classification, IoU prediction regression, bounding box regression, and direction classification, respectively.} %
\label{pipeline}
\vspace*{-2mm}
\end{figure*}

\ifx\allfiles\undefined
\end{document}
\fi 
\ifx\allfiles\undefined
\documentclass[letterpaper]{article}
\begin{document}
\else
\chapter{framework}
\fi

\section{3\quad Confident IoU-Aware Single Stage Detector}
Figure~\ref{pipeline} shows our model's pipeline, which has three parts:
(i) the sparse convolutional network (SPConvNet) for encoding the input point cloud;
(ii) the SSFA module for extracting robust spatial-semantic features; and
(iii) the multi-task head with a confidence function for rectifying the classification score and DI-NMS for post-processing.


\subsection{3.1\quad Point Cloud Encoder}
To encode a point cloud, we voxelize it and calculate the mean coordinates and intensities of points in each voxel as the initial feature.
We then utilize the sparse convolutional network SPConvNet (see Figure~\ref{pipeline}(b)), following the settings of SECOND~\cite{yan2018second} to extract features from the sparse voxels.
The SPConvNet consists of four blocks, each comprising several submanifold sparse convolution (SSC)~\cite{graham20183d} layers and one sparse convolution (SC)~\cite{liu2015sparse} layer.
Specifically, our four blocks have \{2, 2, 3, 3\} SSC layers, respectively, and an SC layer appended to the end of each block for 2x downsampling on the 3D feature maps.
Lastly, we transform the sparse voxel features to dense feature maps and concatenate the features in $z$ to produce the BEV feature maps as inputs to the SSFA module.


\subsection{3.2\quad Spatial-Semantic Feature Aggregation}
To detect cars in autonomous driving, we have to (i) regress precise car locations and also to (ii) classify each regressed bounding box as a positive/negative sample.
In such processes, it is crucial to consider both {\em low-level spatial features\/} and {\em high-level abstract semantic features\/}.
However, when we enrich the high-level abstract semantics in the feature maps, e.g., through stacked convolution layers, the quality of the low-level spatial information often declines as a result.
Hence, the commonly-used BEV feature extraction module~\cite{yan2018second,he2020structure}, which applies stacked convolution layers, could not effectively obtain robust features with rich spatial information.

In this work, we design the {\em spatial-semantic feature aggregation\/} (SSFA) module.
As shown in Figures~\ref{pipeline} (c) \& (d), our SSFA module contains two groups of convolution layers and an attentional fusion module at the end.
The two convolution groups are named the {\em spatial\/} and {\em semantic\/} groups (with corresponding layer-wise features), indicated by the red and green blocks in Figure~\ref{pipeline}(c), respectively, and their outputs are named the spatial and semantic features, respectively.
Specifically, we keep the dimensions (number of channels and feature map size) of the spatial feature to be the same as the input to avoid loss of spatial information.
For the semantic group, we aim to gain more high-level abstract semantic information by taking the spatial feature as input, doubling the number of channels, and reducing the spatial size by half with an initial convolution layer of stride two.
Also, we adopt a 2D DeConv layer to recover the dimensions of the semantic feature map to be the same as the spatial feature map before the element-wise addition to producing the enriched spatial feature (the blue block in Figure~\ref{pipeline}(c)).
On the other hand, we use another 2D DeConv layer to produce the upsampled semantic feature (the yellow block in Figure~\ref{pipeline}(c)) before the attentional fusion.

To adaptively fuse the enriched spatial feature and the upsampled semantic feature, we adopt the attention module shown in Figure~\ref{pipeline}(d).
First, we compress the channels of each feature to one and concatenate the results.
We then use Softmax to normalize the two concatenated channels and split them into two BEV attention maps, in which Softmax builds the dependence between the two features for adaptive feature fusion.
Lastly, we take the learned BEV attention maps as weights on the respective features and perform an element-wise addition ($\oplus$ in Figure~\ref{pipeline}(d)) to fuse the weighted features.
The SSFA module helps extract more robust features with rich spatial and semantic information for more accurate predictions of bounding boxes and classification confidence (see the ablation study in Section~{\color{red} 4.3}).


\begin{figure}
\centering
\includegraphics[width=7.5cm]{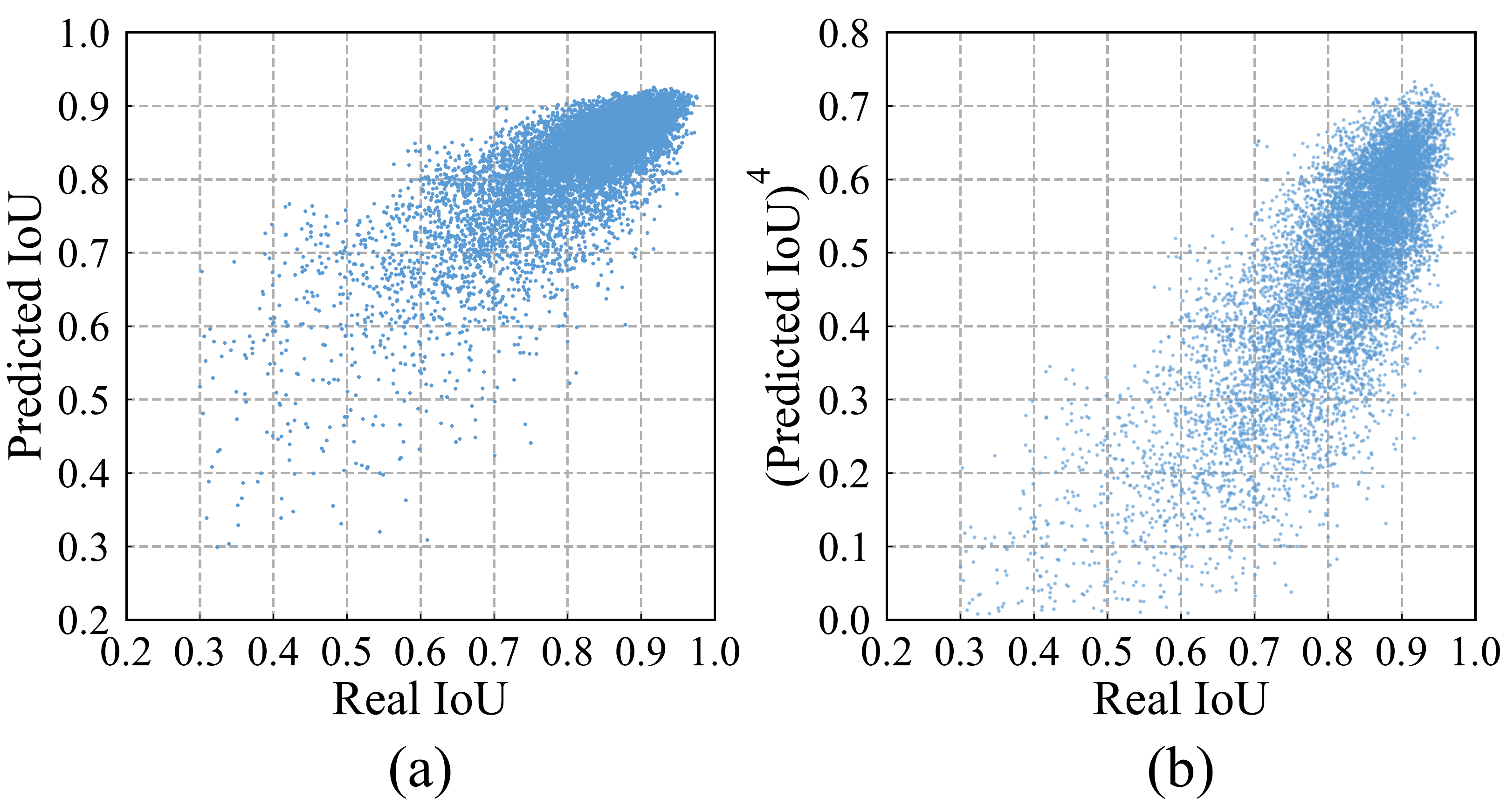}
\vspace*{-3mm}
\caption{Scatterplots:
(a) real IoUs vs. predicted IoUs, showing that high predicted IoU often associates with high real IoU; and
(b) real IoUs vs. (predicted IoUs)$^{\beta}$ with $\beta=4$, such that the rectified IoUs become more discriminative for us to locate good IoU predictions with high certainty.}
\label{fig:stat_iou_pred}
\vspace*{-2mm}
\end{figure}

\subsection{3.3\quad IoU-Aware Confidence Rectification}

To alleviate the misalignment between the localization accuracy and classification confidence without having an additional network stage, we design the {\em IoU-aware confidence rectification module\/} for post-processing the confidence.
In 3D object detection, we usually define anchors by evenly distributing bounding boxes of fixed size on the BEV feature maps; then, we can train the network by regressing the offsets between the anchors and ground truths.
Suffering from the misaligned features of the anchors, the IoUs predicted by anchor-based single-stage detectors are often not as accurate as the IoUs predicted with region proposals in two-stage detectors.
However, we observe from an experiment that the IoUs predicted with anchors are still rather discriminative.

Figure~\ref{fig:stat_iou_pred} shows the experimental result of IoU prediction (``iou reg'' in Figure~\ref{pipeline}(e)) conducted on the KITTI validation set.
Specifically, ``real IoU'' refers to the IoU computed between the anchor-based predicted bounding box and its nearest ground-truth box, whereas ``predicted IoU'' refers to the network-predicted IoU on the corresponding anchor-based predicted bounding box.
From the scatterplot shown in Figure~\ref{fig:stat_iou_pred}(a), we can see that although the predicted IoU cannot perfectly match the real one and the prediction deviation increases when the real IoU drops, high predicted IoUs often associate with high real IoUs, thus allowing us to differentiate between the precise and imprecise bounding box regressions.
Typically, if an anchor feature leads to precise regression, the feature should also predict high IoU with high certainty, as the feature already contains sufficient location information.
On the other hand, if an anchor feature produces imprecise regression, e.g., when the anchor is far away from the ground truths, the feature will likely lead to low-IoU prediction with high uncertainty.

\begin{algorithm}[!t]
    \caption{Distance-variant IoU-weighted NMS}
    \label{alg:dinms}
    \begin{algorithmic}[1]
        \REQUIRE ~~\\
        $\mathcal{C}$ is the $N\times$7 matrix of predicted bounding box parameters $(x, y, z, w, l, h, r)$, where $N$ is the number of bounding boxes, $xyz$ denote box center, $wlh$ denote box dimensions, and $r$ denotes box orientation angle;
        \\
        $\mathcal{S}$ is the set of $N$ rectified confidence values of the corresponding predicted bounding boxes;
        \\
        $\mathcal{I}$ is the set of $N$ IoU prediction values of the corresponding predicted bounding boxes;
        \\
        $\mathcal{A}$ is the $N\times$7 matrix of corresponding anchors for the predicted box parameters $(x, y, z, w, l, h, r)$;
        \\
        $\mathcal{C}=\{c_1,c_2,...,c_N\}; \; \mathcal{I}=\{iou_1,iou_2,...,iou_N\}$;
        \\
        $\mathcal{S}=\{s_1,s_2,...,s_N\}; \; and \; \ \mathcal{A}=\{a_1,a_2,...,a_N\}$.
        \ENSURE ~~\\
        Selected bounding boxes: $\mathcal{B} = \emptyset, \mathcal{L} = \{1,2,...,N\}$
        \\
        \STATE $dist=\{\|c_{i, [0,1]} - a_{i, [0,1]}\|_2\;|\; c_i\in \mathcal{C}, a_i\in \mathcal{A}, i\in \mathcal{L}\}$
        \\
        \STATE $\mathcal{S} = \{s_i\cdot(1-softmax_i(dist)) \;|\; s_i\in \mathcal{S}, i \in \mathcal{L}  \}$
        \\
        \STATE $iou\_thres=0.3, cnt\_thres=\mu$
        \\
        \WHILE{$\mathcal{L} \neq \emptyset$}
        \STATE $cnt=0, idx=\mathop{\arg\max}\limits_{i \in \mathcal{L}}\mathcal{S},c'=c_{idx}$
        \STATE $\mathcal{L}'=\{i\;|\;i \in \mathcal{L}, IoU(c_i,c')>iou\_thres, c_i\in \mathcal{C}\}$
        \STATE $cnt= \mathop{\Sigma}\limits_{i\in \mathcal{L}'} iou_i\cdot IoU(c_i,c'),  c_i\in \mathcal{C}, iou_i\in \mathcal{I}$
        \STATE $b=\frac{\mathop{\Sigma}\limits_{i\in \mathcal{L}'} iou_i\cdot c_i\cdot e^{-(1-IoU(c_i,c'))^2/\sigma^2}}{\mathop{\Sigma}\limits_{i\in \mathcal{L}'} iou_i\cdot e^{-(1-IoU(c_i,c'))^2/\sigma^2}}, c_i\in \mathcal{C}, iou_i\in \mathcal{I}$
        \\
        $\sigma \propto \|c_{i,[0,1]}\|_2$
        \STATE $\mathcal{L}\leftarrow \mathcal{L}-\mathcal{L}'$
        \STATE if $cnt>cnt\_thres$, $\mathcal{B}\leftarrow \mathcal{B}\cup \{b\}$
        \ENDWHILE
        \RETURN $\mathcal{B}$
    \end{algorithmic}
\end{algorithm}

To suppress the uncertainties of low-IoU predictions and further augment the discrimination between low-IoU and high-IoU predictions, we introduce the rectification item $g$:
\begin{equation}\label{ci}
  \begin{matrix}
    g = i^\beta \ , \\
  \end{matrix}
\end{equation}
where $i$ denotes the predicted IoU and $\beta$ is a hyperparameter that controls the degrees of suppressing the low-IoU predictions and augmenting the high-IoU predictions.
As shown in Figure~\ref{fig:stat_iou_pred}(b), the predictions of high IoUs can become more discriminative,~\eg, when setting $\beta=4$.
Hence, we propose to rectify the classification score with $g$ and formulate the following Confidence Function $f$ in the multi-task head:
\begin{equation}
\label{eq:cf}
  \begin{matrix}
    f = c\cdot g = {c} \cdot {i^\beta} \\
  \end{matrix}
\end{equation}
where $c$ is the classification score of the predicted bounding box.
By this means, we can polarize the effect of low-IoU and high-IoU predictions for better rectification of $c$.

During the training, we train the IoU prediction branch simultaneously with the bounding box regression and classification branches and detach the predicted bounding boxes from the computation graph in the IoU predictions to avoid the gradients to back-propagate from the IoU prediction loss to the bounding box regressions.
Only in the testing, we make use of the Confidence Function in Eq.~\eqref{eq:cf} to rectify the classification score and produce the rectified classification confidence $f$ for use in the following NMS process.


\begin{figure*}
\centering
\includegraphics[width=17.3cm]{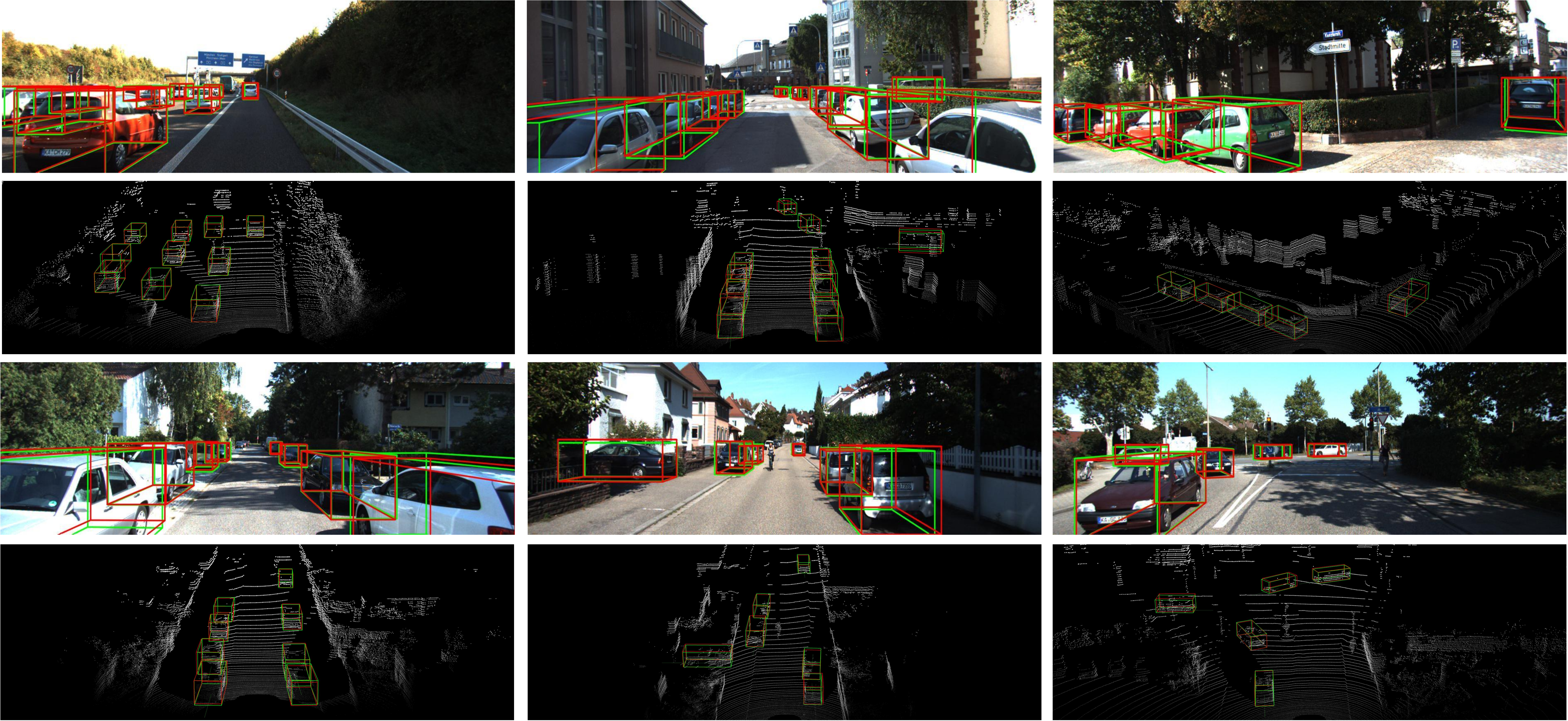}
\vspace*{-2mm}
\caption{Snapshots of our 3D detection results on the KITTI validation set.
The predicted and ground-truth bounding boxes are shown in green and red, respectively, and are projected back onto the color images (1$^{\mathrm{st}}$ \& 3$^{\mathrm{rd}}$ rows) for visualization.}
\label{detresults}
\vspace*{-2mm}
\end{figure*}


\subsection{3.4\quad Distance-Variant IoU-Weighted NMS}
Distant objects are often predicted with low classification confidence and high regression uncertainties, caused by the point sparsity at distant regions.
Hence, we often observe (i) strong oscillations on the regressed bounding boxes and (ii) redundant false-positive predictions that do not overlap with any ground-truth bounding box (in which a few anchors are incorrectly activated to perform the regressions).

To avoid these issues, we formulate the {\em distance-variant IoU-weighted NMS\/} (DI-NMS) for post-processing the predictions.
Algorithm~\ref{alg:dinms} outlines the procedure:
(i) In Steps 1-2, we calculate the $L_2$ distance between the BEV centers of each pair of anchor and predicted box, and take the distance offset to refine the confidence;
(ii) In Steps 5-7, we select box $c'$ with the highest classification confidence and find indices of its overlapped boxes $\mathcal{L'}$ filtered with an IoU threshold, in which $c'$ and $\mathcal{L'}$ are the candidate box and indices of auxiliary boxes, respectively.
Next, we calculate a cumulative sum of products between the IoU predictions of the auxiliary boxes and their real IoUs with the candidate box, in which we design the cumulative sum to filter redundant (zero-IoU) false positives; and
(iii) In Step 8, we use a Gaussian weighted average to obtain a smooth regression from the auxiliary boxes.

Note, auxiliary boxes that have large IoU predictions or high IoUs with the candidate box are assigned with large Gaussian weights.
Also, considering that the localization uncertainty increases with distance, we utilize $\sigma$ to adjust the smoothness degree of the Gaussian weights, in which $\sigma$ is positively correlated with the BEV distance between the box center and viewpoint.
Thus, we can use the smooth Gaussian weights to evenly consider the oscillated auxiliary boxes in distant predictions, while using differentiated weights to focus on the auxiliary boxes that are highly overlapped with the candidate box in short-range predictions.


\subsection{3.5\quad Loss Function}
For model optimization, we follow the conventional settings of box encoding and loss functions in~\cite{yan2018second}.
Specifically, we use the Focal loss~\cite{lin2017focal}, Smooth-$L_1$ loss~\cite{liu2016ssd}, and cross-entropy loss in the bounding box classification ($\mathcal{L}_\text{cls}$), box regression ($\mathcal{L}_\text{box}$), and direction classification ($\mathcal{L}_\text{dir}$).
Besides, we encode the IoUs between the predicted boxes and the ground-truth boxes to keep them in the range of [-1, 1]:
\begin{equation}\label{iouencode}
  \begin{matrix}
    iou_t = {2\cdot(iou - 0.5)} \\
  \end{matrix}
\end{equation}
where $iou$ denotes the real IoU between the predicted bounding box and ground-truth box, and $iou_t$ denotes the encoded target for the IoU prediction.
Then, we apply the Smooth-$L_1$ loss on the IoU regressions to calculate the IoU prediction loss $\mathcal{L}_\text{iou}$.
Similar to the bounding box regression loss $\mathcal{L}_\text{box}$, we calculate the IoU prediction loss $\mathcal{L}_\text{iou}$ only on the positive samples.
The overall loss $\mathcal{L}$ is defined as
\begin{equation}\label{totalloss}
  \begin{matrix}
  $$\mathcal{L}=\mathcal{L}_\text{cls}+\omega\mathcal{L}_\text{box}+\mu\mathcal{L}_\text{dir}+\lambda\mathcal{L}_\text{iou}$$ \\
  \end{matrix}
\end{equation}
where we empirically set $\omega = 2.0$, $\mu = 0.2$, and $\lambda = 1.0$.

\ifx\allfiles\undefined
\end{document}
\fi 
\ifx\allfiles\undefined

\documentclass[letterpaper]{article}
\begin{document}
\else
\chapter{experiments}
\fi

\section{4\quad Experiments}

We evaluate our CIA-SSD on the KITTI 3D object benchmark~\cite{geiger2013vision} with 7,481 training samples and 7,518 test samples.
The training samples are further divided into a training set (3,712 samples) and a validation set (3,769 samples).
Following previous works,~\eg, SASSD~\cite{he2020structure} and SECOND~\cite{yan2018second}, we conducted experiments on the most commonly-used car category and evaluated the results by average precision (AP) with IoU threshold 0.7.
Also, the dataset has three difficulty levels (easy, moderate, and hard) based on the object size, occlusion, and truncation levels.
Figure~\ref{detresults} shows some of our predicted bounding boxes projected onto color images.

\begin{table}[t]
   \centering
   \footnotesize
   \resizebox{\columnwidth}{!}{
   \begin{tabular}{|c|c|c||c|c|c|}
       \hline
       \multicolumn{1}{|c|}{ \multirow{2}{*}{Type}} & \multicolumn{1}{c|}{ \multirow{2}{*}{Method}} & \multicolumn{1}{c||}{ \multirow{2}{*}{Data}} & \multicolumn{3}{|c|}{$AP_{3D} (\%)$} \\ \cline{4-6}
       \multicolumn{1}{|c|}{} & \multicolumn{1}{c|}{} & \multicolumn{1}{c||}{} & \multicolumn{1}{|c|}{Easy} & \multicolumn{1}{|c|}{Mod} & \multicolumn{1}{|c|}{Hard} \\
       \hline
       \hline
      \multirow {12}{*}{2-stage}
         & MV3D~\shortcite{MV3D}                               & {LiDAR+RGB} & 74.97  & 63.63  & 54.00 \\
      {} & F-PointNet~\shortcite{FPOINTNET}                    & {LiDAR+RGB} & 82.19  & 69.79  & 60.59 \\
      {} & AVOD~\shortcite{AVOD}                               & {LiDAR+RGB} & 83.07  & 71.76  & 65.73 \\
      {} & PI-RCNN~\shortcite{xie2020pi}                       & {LiDAR+RGB} & 84.37  & 74.82  & 70.03 \\
      {} & PointRCNN~\shortcite{shi2019pointrcnn}              & {LiDAR}     & 86.96  & 75.64  & 70.70 \\
      {} & F-ConvNet~\shortcite{wang2019frustum}               & {LiDAR+RGB} & 87.36  & 76.39  & 66.69 \\
      {} & 3D IoU Loss~\shortcite{zhou2019iou}                 & {LiDAR}     & 86.16  & 76.50  & 71.39 \\
      {} & Patches~\shortcite{lehner2019patch}                 & {LiDAR}     & 88.67  & 77.20  & 71.82 \\
      {} & Fast PointRCNN~\shortcite{Chen2019fastpointrcnn}    & {LiDAR}     & 85.29  & 77.40  & 70.24 \\
      {} & UberATG-MMF~\shortcite{liang2019multi}              & {LiDAR+RGB} & 88.40  & 77.43  & 70.22 \\
      {} & Part-$A^2$~\shortcite{shi2020points}                & {LiDAR}     & 87.81  & 78.49  & 73.51 \\
      {} & 3D IoU-Net~\shortcite{li20203d}                     & {LiDAR}     & 87.96  & 79.03  & 72.78 \\
      {} & STD~\shortcite{yang2019std}                         & {LiDAR}     & 87.95  & 79.71  & 75.09 \\
      {} & 3D-CVF~\shortcite{yoo20203d}                        & {LiDAR+RGB}  & 88.84  & 79.72  & 72.80 \\
      {} & PV-RCNN~\shortcite{shi2020pv}                       & {LiDAR}     & {\em 90.25}  & {\em 81.43}  & {\em 76.82} \\
      \hline
      \hline
      \multirow {11}{*}{1-stage}
       & VoxelNet~\shortcite{zhou2018voxelnet}                 &{LiDAR}      & 77.82  & 64.17  & 57.51 \\
      {} & ContFuse~\shortcite{CONTFUSE}                       &{LiDAR+RGB}  & 83.68  & 68.78  & 61.67 \\
      {} & SECOND~\shortcite{yan2018second}                    &{LiDAR}      & 83.34  & 72.55  & 65.82 \\
      {} & PointPillars~\shortcite{lang2019pointpillars}       &{LiDAR}      & 82.58  & 74.31  & 68.99 \\
      {} & TANet~\shortcite{liu2020tanet}                      &{LiDAR}      & 84.39  & 75.94  & 68.82 \\
      {} & Associate-3Ddet~\shortcite{du2020associate}         &{LiDAR}      & 85.99  & 77.40  & 70.53 \\
      {} & Point-GNN~\shortcite{shi2020point}                  &{LiDAR}      & 88.33  & 79.47  & 72.29 \\
      {} & 3DSSD~\shortcite{yang20203dssd}                     &{LiDAR}      & 88.36  & 79.57  &\bf 74.55 \\
      {} & SASSD~\shortcite{he2020structure}                   &{LiDAR}      & 88.75  & 79.79  & 74.16 \\ \cline{2-6}
      {} & CIA-SSD (ours)                                      &{LiDAR} & \bf 89.59  & \bf 80.28  & 72.87 \\
      \hline
   \end{tabular}
   }
   \vspace*{-2mm}
   \caption{Comparison with the state-of-the-art methods on the KITTI test set.
   The 3D average precisions of 40 sampling recall points for car detection are evaluated on the KITTI official server; from the official ranking metric ``{\em Moderate AP\/},'' we can see that our CIA-SSD attains the top performance compared with all 1-stage detectors and is comparable with the top 2-stage detector on this challenging problem.
   }
   \label{table1}
\end{table}

\begin{table}[t]
   \centering \addtolength{\tabcolsep}{-1pt}
   \footnotesize
   \begin{tabular}{|c|c|c|c|}
      \hline
       \multicolumn{1}{|c|}{ \multirow{2}{*}{Method}} & \multicolumn{3}{|c|}{$AP_{3D} (\%)$} \\ \cline{2-4}
       \multicolumn{1}{|c|}{}  & \multicolumn{1}{|c|}{Easy} & \multicolumn{1}{|c|}{Mod} & \multicolumn{1}{|c|}{Hard} \\
       \hline
       \hline
       VoxelNet~\shortcite{zhou2018voxelnet}          & 81.97  & 65.46 & 62.85 \\
       ContFuse~\shortcite{CONTFUSE}                  & 86.32  & 73.25 & 67.81 \\
       SECOND~\shortcite{yan2018second}               & 87.43  & 76.48 & 69.10 \\
       TANet~\shortcite{liu2020tanet}                 & 88.21  & 77.85 & 75.62  \\
       PointPillars~\shortcite{lang2019pointpillars}  &  -     & 77.98 & - \\
       Point-GNN~\shortcite{shi2020point}             & 87.89  & 78.34 & 77.38 \\
       Associate-3Ddet~\shortcite{du2020associate}    & 89.29  & 79.17 & 77.76 \\
       3DSSD~\shortcite{yang20203dssd}                & 89.71  & 79.45 & 78.67 \\
       SASSD~\shortcite{he2020structure}              & \bf90.15  & \bf79.91 & 78.78 \\ \cline{1-4}
       CIA-SSD (ours)                                 & 90.04  & 79.81 & \bf78.80 \\
      \hline
   \end{tabular}
   \vspace*{-2mm}
   \caption{Comparison with the state-of-the-art single-stage detectors on the KITTI validation set, in which the 3D average precisions for car detection are based on 11 sampling recall points (vs. 40 points in the KITTI test set).}
   \label{table2}
   \vspace{-2mm} 
\end{table}


\subsection{4.1\quad Implementation Details}

We voxelize the input point cloud into a grid of resolutions [0.05, 0.05, 0.1] meters in ranges [0, 70.4], [-40, 40], and [-3, 1] meters along the $x$, $y$, and $z$ axes, respectively.
The anchors are pre-defined evenly on the BEV feature map with the same dimensions (width=1.6m, length=3.9m, height=1.56m) and two possible orientations (0{\degree} or 90{\degree}).
Further, they are divided into three categories (positive, negative, and ignored) based on the matching strategy in~\cite{zhou2018voxelnet} with IoU thresholds 0.45 and 0.6.
 
We consider four types of data augmentations to enhance our model's generalization ability.
The first type is a global augmentation on the entire point cloud, including random rotation, scaling, and flipping.
The second type is a local augmentation on a portion of the point cloud around a ground-truth object, including random rotation and translation.
The third type is a ground-truth augmentation following SECOND~\cite{yan2018second}.
Last, we filter out objects with difficulty levels not attributed to easy, moderate, and hard to improve the quality of the positive samples, and take also objects of similar categories, such as van for car, as the targets to alleviate model confusion in the training.

In the SSFA module, the spatial and semantic groups have three stacked convolution layers of kernel 3x3 with a number of channels 128 and 256, respectively.
After the spatial and semantic groups, there is one 1x1 convolution layer with 128 and 256 channels separately.
The 2D DeConv layers have 3x3 kernels and 128 output channels with stride two.
Before the attentional fusion, we use a 3x3 convolution layer with 128 output channels to transform each group feature.
In the attentional fusion, the convolutional layers have 3x3 kernels and one output channel.
Other network settings follow those in SECOND~\cite{yan2018second}.

We use the ADAM optimizer~\cite{kingma2014adam} with the cosine annealing learning rate~\cite{loshchilov2016sgdr} to train our model with a batch size of four on a single GPU card for 60 epochs.
Further, we empirically set $\beta\;$=\;4 (in the confidence function), $\mu\;$=\;2.6 (in DI-NMS), and $\sigma\;$=\;\{0.0009, 0.009, 0.1, 1\} (in DI-NMS) for BEV distances in ranges [0, 20m), [20m, 40m), [40m, 60m), and [60m, 70.4m], respectively.


\subsection{4.2\quad Comparison with State-of-the-Arts}
We compare our CIA-SSD with the state-of-the-art methods listed in Table~\ref{table1}.
As shown in the table, our model ranks the 1$^{st}$ place in terms of moderate and easy AP,~\ie, 80.28\% and 89.59\% respectively, among all the single-stage detectors.
The ``moderate AP'' is the official ranking metric for 3D detection on the KITTI official test server.
Our CIA-SSD outperforms all the state-of-the-art single-stage detectors, including the very recent ones, including Point-GNN, 3DSSD, and SASSD by about 0.5 to 0.8 points under this metric.
Besides, while two-stage detectors generally perform better than single-stage detectors due to the extra second-stage refinement,
our proposed single-stage detector still outperforms most of the recent two-stage detectors,~\eg, 3D-CVF, STD, and Part-$A^2$ by about 0.6 to 1.8 points on moderate AP.
Furthermore, we shall show the high efficiency of our model compared with two-stage detectors in Section {\color{red} 4.4}.

Although our model sets the new state-of-the-art single-stage results on easy and moderate APs for the KITTI test set, the corresponding APs are slightly lower than some of the state-of-the-art results on the validation set, as shown in Table~\ref{table2}.
Also, our hard AP is lower than the state-of-the-art methods on the test set, while being the top on the validation set.
We argue that such inconsistency may be caused by the mismatched distributions between the KITTI val and test splits, as mentioned in Part-$A^2$~\cite{shi2020points}.


\subsection{4.3\quad Ablation Study}

We adopt the 2D feature extraction module~\cite{yan2018second} grouped by seven stacked convolutional layers, the normal NMS, and multi-task head without the IoU prediction branch as the baseline modules in the ablation study.

\begin{table}[!t]
    \small
    \begin{center}
        \scalebox{1.0}{
            \begin{tabular}{ccccc|ccc}
                \hline
                \tabincell{c}{\textit{glo.}}
                & \tabincell{c}{\textit{loc.}}
                & \tabincell{c}{\textit{gt aug.}}
                & \tabincell{c}{\textit{sim.}}
                & \tabincell{c}{\textit{diff.}}
                & Easy & Mod & Hard \\
                \hline
                \hline
                           &            &            &            &            & 81.29 & 66.39 & 65.40 \\
                \checkmark &            &            &            &            & 86.53 & 75.73 & 74.77 \\
                \checkmark & \checkmark &            &            &            & 87.18 & 76.33 & 68.64 \\
                \checkmark & \checkmark & \checkmark &            &            & 87.97 & 78.03 & 76.80 \\
                \checkmark & \checkmark & \checkmark &\checkmark  &            & 88.81 & 78.35 & 77.26 \\
                \checkmark & \checkmark & \checkmark &\checkmark  &\checkmark  & \bf89.09 & \bf78.73 & \bf77.56 \\
                \hline
            \end{tabular}
        }
    \end{center}
    \vspace*{-2mm}
    \caption{Ablation study on our implemented data processing techniques on the baseline modules, in which we report the 3D average precisions of 11 sampling recall points for car detection on the KITTI val split.
    Here, ``glo.,'' ``loc,'' ``gt aug.,'' ``sim.,'' and ``diff.'' denote the global augmentation, local augmentation, ground truth augmentation, training with similar type of objects, and filtering objects with difficulty levels not attributed to easy, moderate, \& hard, respectively.}
    \label{table3}
\end{table}

\begin{table}[!t]
    \small
    \begin{center}
        \scalebox{1.0}{
            \begin{tabular}{ccc|ccc}
                \hline
                \tabincell{c}{\textit{SSFA}}
                & \tabincell{c}{\textit{CF}}
                & \tabincell{c}{\textit{DI-NMS}}
                & Easy & Mod  & Hard \\
                \hline
                \hline
                           &            &             & 89.09 & 78.73 & 77.56 \\
                \checkmark &            &             & 89.46 & 79.17 & 77.88 \\
                \checkmark & \checkmark &             & 89.66 & 79.63 & 78.64 \\
                \checkmark & \checkmark & \checkmark  & \bf90.04 & \bf79.81 & \bf78.80 \\
                \hline
            \end{tabular}
        }
    \end{center}
    \vspace*{-2mm}
    \caption{Ablation study on our modules: SSFA, CF, and DI-NMS.
    Here, we report the 3D average precisions of 11 sampling recall points for car detection on the KITTI val split.}
    \label{table4}
\end{table}

{\bf Effect of data processing}
Table~\ref{table3} shows results that reveal the effect of each data processing technique on the baseline modules.
Both global and local augmentations effectively improve easy and moderate APs.
Also, the ground truth augmentation, training with the similar type of objects, and
filtering objects with suitable difficulty levels for ground-truth augmentation boost all levels of AP effectively.
All these techniques help to build up a strong baseline for better validation of our proposed modules.

{\bf Effect of SSFA module}
As shown in Table~\ref{table4}, our SSFA module contributes an improvement of 0.37, 0.44, and 0.32 on the moderate, easy, and hard APs, respectively.
Besides, compared with the baseline 2D feature extraction module under a batch size of four, our SSFA module and the corresponding module in SASSD~\cite{he2020structure} increase our framework's GPU occupation by about 10\% and 27\%, respectively, validating that our SSFA module is lightweight.

{\bf Effect of confidence function}
In Table~\ref{table4}, our confidence function in the multi-task head improves the easy, moderate, and hard APs by 0.20, 0.46, and 0.76, respectively.
Also, we use the metric Pearson Correlation Coefficient (PCC)~\cite{pearson} to measure the correlation between the IoU and confidence of our predicted boxes and calculate the corresponding moderate AP for different $\beta$ values in Table~\ref{table5}.
We can see that setting $\beta\;$=\;4 leads to the highest PCC and moderate AP.
Besides, the AP still increases from 79.17 to 79.30 when setting $\beta\;$=\;0 (\ie, without rectifying the classification scores), because the IoU prediction optimizes the model to make the learned features aware of the relative locations between the predicted boxes and ground truths.

{\bf Effect of DI-NMS}
Our DI-NMS raises the easy AP (by 0.38), moderate AP (by 0.18), and hard AP (by 0.16); see Table~\ref{table4}.
However, different from 2D detection with region proposals that often fully cover the objects,
candidate boxes in 3D detection may not have good object coverage, due to the missing points around the boundary, so it is hard to produce well-aligned boxes with the weighted average in DI-NMS.
Hence, the increase in AP with DI-NMS is lower than that with the SSFA module and confidence function.

\begin{table}[!th]
    \small
    \begin{center}
    \resizebox{\columnwidth}{!}{
        \scalebox{1.0}{
            \begin{tabular}{cc@{\hspace{3mm}}c@{\hspace{3mm}}c@{\hspace{3mm}}c@{\hspace{3mm}}c@{\hspace{3mm}}c@{\hspace{3mm}}c}
                \hline
                \multicolumn{1}{c|}{$\beta$}
                & 0      & 1     & 2     & 3     & 4     & 5  & 6 \\
                \hline
                \multicolumn{1}{c|}{AP$_{\text{Mod}}$}
                & 79.30   & 79.37   & 79.56  &79.61  & \bf79.63  &79.62  &79.61
                \\
                \multicolumn{1}{c|}{PCC}
                & 0.460
                & 0.471
                & 0.502
                & 0.509
                & \bf0.511
                & 0.510
                & 0.509 \\
                \hline
            \end{tabular}
        }}
    \end{center}
    \vspace*{-2mm}
    \caption{Hyperparameter analysis on $\beta$ in the confidence function.
    Here, we report the 3D moderate average precision (AP$_{\text{Mod}}$) and Pearson Correlation Coefficient (PCC) between the IoU and confidence of the predicted boxes for different $\beta$ values in car detection on the KITTI val split.}
    \label{table5}
\end{table}

\begin{table}[!th]
    \small
    \begin{center}
    \resizebox{\columnwidth}{!}{
        \scalebox{1.0}{
            \begin{tabular}{cc@{\hspace{3mm}}c@{\hspace{3mm}}c@{\hspace{3mm}}ccc}
                \hline
                \multicolumn{1}{c|}{1-stage}    & Point-GNN &Associate-3Ddet & SASSD & 3DSSD & TANet     & \multicolumn{1}{|c}{Ours (1-stage)}   \\
                \hline
                \multicolumn{1}{c|}{time (ms)}  & 643       &60 & 40.1  & 38    & 34.75     & \multicolumn{1}{|c}{\bf30.76}  \\
                \hline
            \end{tabular}
        }}
    \end{center}
    \vspace*{-2.5mm}
    \caption{Comparing the runtime (in millisecond) of our model with very recent state-of-the-art single-stage detectors, showing that our model runs the fastest among them.}
    \label{table7}
\end{table}

\begin{table}[!th]
    \small
    \begin{center}
    \resizebox{\columnwidth}{!}{
        \scalebox{1.0}{
            \begin{tabular}{cc@{\hspace{3mm}}c@{\hspace{3mm}}ccccc}
                \hline
                \multicolumn{1}{c|}{2-stage}         &  PointRCNN & Part-$A^2$ & STD    & Fast PointRCNN   & \multicolumn{1}{|c}{Ours (1-stage)} \\
                \hline
                \multicolumn{1}{c|}{time (ms)}  & 100         & 80   & 80       & 65    & \multicolumn{1}{|c}{\bf30.76}\\
                \hline
            \end{tabular}
        }}
    \end{center}
    \vspace*{-2.5mm}
    \caption{Comparing the runtime (in millisecond) of our model with state-of-the-art two-stage detectors, showing that our single-stage detector has much higher efficiency.}
    \label{table8}
    \vspace*{-1mm}
\end{table}

\subsection{4.4\quad Runtime Analysis}
Table~\ref{table7} compares the runtime of our CIA-SSD with five very recent state-of-the-art single-stage detectors, and we can see that our CIA-SSD is the fastest one.
Notice that our CIA-SSD is faster than SASSD as the number of channels in our SSFA module is only half of that of SA-SSD in most layers.
Next, Table~\ref{table8} compares the runtime of CIA-SSD with four recent two-stage detectors.
From the table, we can see that CIA-SSD runs significantly faster than these two-stage detectors, confirming the high runtime efficiency of single-stage detectors.
The average inference time of CIA-SSD is 30.76ms, including
(i) 2.84ms for data processing before the network forwarding;
(ii) 24.33ms for data processing in the network; and
(iii) 3.59ms for post-processing to produce the final predictions.
All the reported timing results were averaged from five runs of our program on an Intel Xeon Silver CPU and a single TITAN Xp GPU.


\section{5\quad Conclusion}

This paper presents a new object detector on point clouds, named Confident IoU-Aware Single-Stage object Detector (CIA-SSD).
Our main contributions include the spatial-semantic feature aggregation module for extracting robust spatial-semantic features for object predictions, the formulation of a confidence function to rectify the classification score and alleviate the misalignment between the localization accuracy and classification confidence, and the distance-variant IoU-weighted NMS to obtain smoother results and avoid redundant (zero-IoU) false positives.
The experimental results show that CIA-SSD achieves the state-of-the-art 3D detection performance on the official ranking metric (Moderate AP) for the KITTI benchmark, compared with all the existing single-stage detectors.
Also, CIA-SSD attains real-time detection efficiency and runs the fastest, compared with the very recent state-of-the-art detectors.

\section{Acknowledgement}
We thank reviewers for the valuable comments. This work is funded by the Research Grants Council of the Hong Kong Special Administrative Region (Proj. no. CUHK 14201918).

\bibliography{mybib}
\bibliographystyle{aaai21}

\ifx\allfiles\undefined
\end{document}
\fi

\end{document}